\documentclass[submission,copyright,creativecommons]{eptcs}
\providecommand{\event}{End-to-End Compositional Models of Vector-Based Semantics} 

\usepackage{breakurl}             
\usepackage{underscore}           

\usepackage{float}
\usepackage[shortlabels]{enumitem}
\usepackage[table,xcdraw]{xcolor}
\usepackage[linguistics]{forest}
\usepackage{tikz}
\usetikzlibrary{decorations.pathreplacing,calligraphy}
\usepackage{diagbox}
\usepackage{amsmath}
\usepackage{subcaption}
\usepackage{multicol}
\usepackage{wrapfig}
\usepackage[T1]{tipa}
\usepackage{substitutefont}
\usepackage{multirow}

\title{Word-Embeddings Distinguish Denominal and Root-Derived Verbs in Semitic}
\author{Ido Benbaji\thanks{All authors contributed equally.}
\institute{MIT\\
Cambridge, MA, USA}
\email{ibenbaji@mit.edu}
\and
Omri Doron$^*$
\institute{MIT\\
Cambridge, MA, USA}
\email{omrid@mit.edu}
\and
Adèle Hénot-Mortier$^*$
\institute{MIT\\
Cambridge, MA, USA}
\email{mortier@mit.edu}
}

\def\titlerunning{Word-Embeddings Distinguish Denominal and Root-Derived Verbs in Semitic}
\def\authorrunning{I. Benbaji, O. Doron \& A. Hénot-Mortier}

\begin{document}
\sloppy
\maketitle
\begin{abstract}
    Proponents of the Distributed Morphology framework have posited the existence of two levels of morphological word formation: a lower one, leading to loose input-output semantic relationships; and an upper one, leading to tight input-output semantic relationships. In this work, we propose to test the validity of this assumption in the context of Hebrew word embeddings. If the two-level hypothesis is borne out, we expect state-of-the-art Hebrew word embeddings to encode (1) a noun, (2) a denominal derived from it (\textit{via} an upper-level operation), and (3) a verb related to the noun (\textit{via} a lower-level operation on the noun's root), in such a way that the denominal (2) should be closer in the embedding space to the noun (1) than the related verb (3) is to the same noun (1). We report that this hypothesis is verified by four embedding models of Hebrew: fastText, GloVe, Word2Vec and AlephBERT. This suggests that word embedding models are able to capture complex and fine-grained semantic properties that are morphologically motivated. 
\end{abstract}

\section{Introduction: A few basic principles of word formation}
\subsection{Morphological processes sometimes appear ``irregular''}
A common assumption in generative morphology is that word formation differs essentially from the formation of larger phrasal structures \cite{Postal1969, Chomsky1970, Chomsky1973}. Phrase formation on the one hand, is generally productive in the sense that it is seldom subject to arbitrary constraints; it also proves to be semantically compositional. Word formation on the other hand, is compositional at times and non-compositional at others, and is known to exhibit arbitrary paradigmatic gaps \cite{Aronoff1976}.

The lack of compositionality in word formation as been observed in certain English ``berry-words'' \cite{Hervey1973, Aronoff1976}. The berry-words in (1a) seem semantically compositional as they involve concatenation of two independent morphemes, each of which contributes its meaning to the word as a whole. In contrast, the berry-words in (1b) cannot be said to be compositional, as one of their morphemes fails to convey any meaning when uttered independently from the other (i.e., the morphemes \textit{cran, boysen,} and \textit{huckle} are not meaningful units in English).
\begin{enumerate}[(1)]
	\item \begin{enumerate}[a.]
		\item \textit{\textbf{black}berry, \textbf{blue}berry} \hfill ``compositional'' berries\label{compositional berries}
		\item \textit{\textbf{cran}berry, \textbf{boysen}berry, \textbf{huckle}berry} \hfill ``non-compositional'' berries\label{non-compositional berries}
	\end{enumerate}\label{ity affixation}
\end{enumerate}

For an example of the arbitrariness in the application of morphological processes, consider the English nominals in (2). (2a) illustrates a regular morphological process that merges the morpheme \textit{-ity} to an adjective ending with the phonemes /\textit{-(i)ous}/, resulting in a noun. As is illustrated in (2b), some arbitrary constraint prevents this process from applying to the adjective \textit{atrocious} \cite{Aronoff1976}. Instead, a different process seems to apply in the derivation of \textit{atrocity} from \textit{atrocious} -- perhaps one which involves truncation of part of the adjective, followed by the merger of the nominalizing morpheme.
\begin{enumerate}[(1), resume]
	\item \begin{enumerate}[a.]
		\item \textit{curious\textbf{ity}, monstros\textbf{ity}, pompos\textbf{ity}} \hfill regular \textit{ity}-nominalization\label{systematic derivations}
		\item \textit{atroc\textbf{ity}. *atrocious\textbf{ity}} \hfill irregular \textit{-ity}-nominalization\label{non-systematic derivations}
	\end{enumerate}
\end{enumerate}

\subsection{The two-level model}
To account for this ambivalent nature of morphological processes, linguists working in the tradition of the \textit{Distributed Morphology} (DM) framework (\cite{Halle1993,Embick2010,Bobaljik2012,Harley2014}) have posited the existence of two ``levels'' of morphological derivation; a ``lower'' level, where word formation may be irregular, arbitrary, and non-productive, and an ``upper'' level, where word formation is mostly regular and productive (cf. \cite{Marantz2000}).

\begin{wrapfigure}{r}{0.35\textwidth}
    \centering
    \vspace{4mm}
    \scalebox{0.9}{
    \begin{forest}, nice empty nodes
        [[$\cdots$][$\cdots$[[$\sqrt{~}$][$\cdots$]][{n, a, v}]]]
        \node[] at (1.2,-1.25) {$\longleftarrow$word};
        \coordinate (low1) at (-1.2,-1.3);
        \coordinate (low2) at (-1.2,-3.5);
        \draw [decorate, decoration = {calligraphic brace}] (low2) --  (low1);
        \coordinate (high1) at (-1.2,0);
        \coordinate (high2) at (-1.2,-1.25);
        \draw [decorate, decoration = {calligraphic brace}] (high2) --  (high1);
        \node[text width=3cm] at (-2.3,-2.4) {``lower'' level};
        \node[text width=3cm] at (-2.3,-0.6) {``upper'' level};
    \end{forest}
    }
    \caption{Basic decomposition of the word formation process}
    \label{fig:word-formation-process}
\end{wrapfigure}
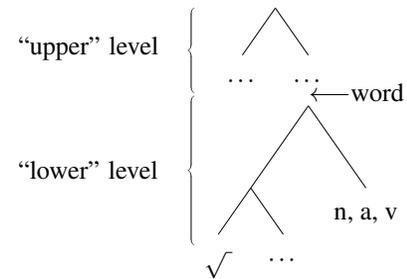

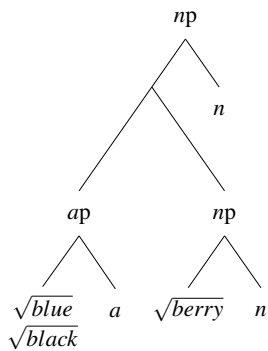
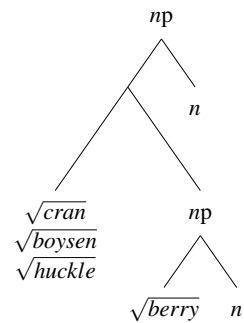
\begin{figure}[H]
    \centering
    \begin{subfigure}[t]{0.45\textwidth}
    \centering
    \scalebox{0.8}{
        \begin{forest}, nice empty nodes
            [\textit{n}p[[\textit{a}p [$\sqrt{blue}$\\$\sqrt{black}$] [\textit{a}]] [\textit{n}p [$\sqrt{berry}$] [\textit{n}]]] [\textit{n}]]
        \end{forest}
    }
        \caption{Compositional case: the 2 sub-words are merged with their respective heads (\textit{a}, \textit{n}) before being compounded.}
    \end{subfigure}\hfill
    \begin{subfigure}[t]{0.45\textwidth}
    \centering
    \scalebox{0.8}{
        \begin{forest}, nice empty nodes
        [\textit{n}p [[$\sqrt{cran}$\\$\sqrt{boysen}$\\$\sqrt{huckle}$] [\textit{n}p [$\sqrt{berry}$] [\textit{n}]]] [\textit{n}]]
        \end{forest}
    }
        \caption{Non-compositional case: 1 of the sub-words is not merged with a functional head before compounding, and thus remains semantically opaque.}
    \end{subfigure}
    \caption{Compositional \textit{vs} non-compositional berry-words}
    \label{fig:berries}
\end{figure}

The DM framework assumes that while morphological processes apply as part of the syntactic derivation, the ``upper'' and ``lower'' levels of morphological derivation are distinguished by the merger of a so-called ``functional head'' (\textit{n}, \textit{v}, \textit{a}, etc.). A functional head sets the semantic, syntactic and phonological features of the word it creates. It may be merged directly with a \textit{root}, which is an atomic element devoid of functional material \cite{Arad2003}; or it may be merged with some other non-atomic constituent already dominating a functional head. More specifically, it has been assumed that any operation that applies directly to a root, (i.e., an operation that applies at the ``lower'', non-word level), should remain impenetrable to upper-level operations. Fig. \ref{fig:berries} below is an illustration of how this model accounts for the difference between the compositional berry-names (\ref{compositional berries}) and the non-compositional ones (\ref{non-compositional berries}). 

\vspace{-5mm}

In the compositional case, the roots of the first morpheme combine with an adjectivizing head (\textit{a}) to form an adjective, and the root of the second morpheme combines with a nominalizing head (\textit{n}) to form a noun. The two are then conjoined to form a word whose meaning is compositional in that it simply involves intersecting the meaning of the adjective with that of the noun. In the non-compositional case, the root that does not convey any meaning in the context of the whole compound ($\sqrt{cran}$, $\sqrt{boysen}$, $\sqrt{huckle}$), is not merged with any functional head before it combines with the noun (\textit{berry}). The derivation is in that case a ``low'' level process. The divide between pre-word ``lower'' level and post-word ``upper'' level processes leads to two related predictions:\\

\indent\textbf{Prediction (a): Elements derived from the same \textit{root} via a lower-level morphological operation may arbitrarily differ semantically;}\vspace{2mm}\\ 
\indent\textbf{Prediction (b): Elements that derive from the same \textit{word} via upper-level morphological operations should be closely related semantically.}

\subsection{The relevance of Semitic morphology}
Semitic languages provide a useful testing ground for the predictions of the two-level model. This is because in many of these languages, words can be decomposed into consonantal roots and fixed morphophonemic patterns that introduce functional information, like part of speech (\textit{n}, \textit{a}, \textit{v} etc.) and valence in the case of verbs (\cite{Tobin2004}). As is illustrated in Tab. \ref{fig:varieties-of-meaning}, patterns can be seen as recipes to ``fill the gaps'' between the consonants of the root. These roots and patterns, which only form independent words when merged together, have nevertheless been argued to be mentally represented as accessible independent morphological units (cf. \cite{Prunet2000}).
\begin{table}[H]
    \centering
    \small
    \begin{tabular}{l|c|cl|cl|}
        \cline{2-6}
        & \backslashbox{Verbal pattern}{Root} & $\sqrt{\text{\textipa{\textbf{xSv}}}}$ &  & $\sqrt{\text{\textipa{\textbf{ktv}}}}$ &  \\ \cline{2-6} (i)&   \textbf{C}a\textbf{C}a\textbf{C} &  \textipa{\textbf{x}a\textbf{S}a\textbf{v}}   & `thought' & \textipa{\textbf{k}a\textbf{t}a\textbf{v}} & `wrote'\\
        (ii)&    ni\textbf{CC}a\textbf{C} & \textipa{ne\textbf{xS}a\textbf{v}} & `was well considered' & \textipa{ni\textbf{xt}a\textbf{v}} & `was written' \\
        (iii)&    \textbf{C}i\textbf{CC}e\textbf{C} & \textipa{\textbf{x}i\textbf{S}e\textbf{v}} & `calculated' & \textipa{\textbf{k}i\textbf{t}e\textbf{v}} & `CC-ed'\\
        (iv)&    \textbf{C}u\textbf{CC}a\textbf{C} & \textipa{\textbf{x}u\textbf{S}a\textbf{v}} & `was calculated' & \textipa{\textbf{k}u\textbf{t}a\textbf{v}} & `was CC-ed' \\
        (v)&    hi\textbf{CC}i\textbf{C} & \textipa{hi\textbf{xS}i\textbf{v}} & `considered' & \textipa{hi\textbf{xt}i\textbf{v}} & `dictated' \\
        (vi)&    hu\textbf{CC}a\textbf{C} & \textipa{hu\textbf{xS}a\textbf{v}} & `was considered' & \textipa{hu\textbf{xt}a\textbf{v}} & `was dictated' \\
        (vii)&    hit\textbf{C}a\textbf{CC}e\textbf{C} & \textipa{hit\textbf{x}a\textbf{S}e\textbf{v}} & `was considerate' & \textipa{hit\textbf{k}a\textbf{t}e\textbf{v}} & `corresponded' \\ \cline{2-6}
    \end{tabular}
    \caption{Varieties of meaning for the same root}
    \label{fig:varieties-of-meaning}
\end{table}

\vspace{-3mm}
While Semitic roots do seem to signal some broad semantic field, the meaning that results from combining a certain root with a certain pattern is highly unpredictable and non-systematic. This is also illustrated in Tab. \ref{fig:varieties-of-meaning}, where two tri-consonantal roots ($\sqrt{\text{\textipa{xSv}}}$ and $\sqrt{\text{\textipa{ktv}}}$), are combined with each of the language's seven verbal patterns. Crucially however, there is no transparent semantic operation that the patterns seem to denote. While niCCaC (ii) seems to constitute the passive version of CaCaC (i) when the two patterns are combined with the root $\sqrt{\text{\textipa{ktv}}}$, this is not the case when they are combined with the root $\sqrt{\text{\textipa{xSv}}}$. Some patterns seem to behave more systematically than others (for instance, huCCaC (vi) is usually just the passive of hiCCiC (v)), but this is not generally the case.

It is also worth noting that Hebrew has three verbal patterns containing four consonant slots -- CiCCeC (iii), CuCCaC (iv), and hitCaCCeC (vii). As pointed out in \cite{Arad2003}, no gemination exists in these patterns in Modern Hebrew. However, we nevertheless have evidence that these patterns contain an extra consonant slot, as they \textit{productively} combine with roots of four consonant. For instance, the root $\sqrt{\text{\textipa{SxKK}}}$ can combine with these patterns to form \textipa{SixKeK} (`liberated'), \textipa{SuxKaK} (`was liberated'), and \textipa{hiStaxKeK} (`liberated oneself'). Other patterns (such as hiCCiC (v)) can combine with four non-templatic consonants, but this process is less productive and is generally restricted to loan-words (for instance, the Hebrew verb \textipa{hiSpKi\texttslig} (`splashed') is derived from the German verb with the same meaning \textit{spritzen}).
This will be relevant to our discussion of the behavior of denominal verbs in Hebrew in the next section. 

\section{The case of Hebrew denominal verbs}
Given our two-level model of morphological processes, root-pattern combination is a lower-level operation by definition. Therefore, it is not surprising that the meanings obtained from combining the same root with different patterns differ in arbitrary ways. Furthermore, a given root can yield forms with potentially divergent syntactic categories, as the root can combine with functional heads of various labels (\textit{n}, \textit{v}, \textit{a}, etc.). This is illustrated in Tab. \ref{fig:nominal-templates}, which combines the same roots used in Tab. \ref{fig:varieties-of-meaning} with nominal and adjectival patterns, rather than verbal ones.

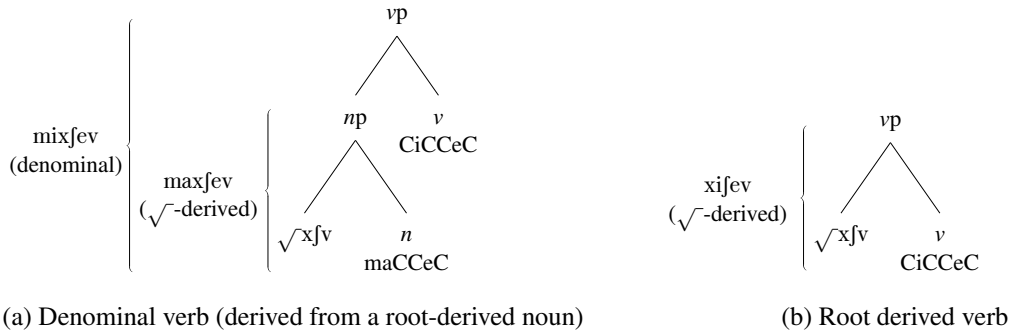
\begin{figure}[H]
    \centering
    \begin{subfigure}[b]{.55\linewidth}
    \scalebox{.8}{
        \begin{forest}, nice empty nodes
            [\textit{v}p [\textit{n}p [$\sqrt{~}$\textipa{xSv}] [\textit{n}\\maCCeC]] [\textit{v}\\CiCCeC]  ]
            \coordinate (low1) at (-2.1,-1.5);
            \coordinate (low2) at (-2.1,-4.2);
            \draw [decorate, decoration = {calligraphic brace}] (low2) --  (low1);
            \coordinate (high1) at (-4.4,0);
            \coordinate (high2) at (-4.4,-4.2);
            \draw [decorate, decoration = {calligraphic brace}] (high2) --  (high1);
            \node[text width=3cm] at (-3.3,-2.8) {\begin{center}
                \textipa{maxSev}\\($\sqrt{~}$-derived)
            \end{center}};
            \node[text width=3cm] at (-5.5,-2) {\begin{center}
                \textipa{mixSev}\\(denominal)
            \end{center}};
        \end{forest}
    }
        \caption{Denominal verb (derived from a root-derived noun)}
        \label{fig:denominal-structure}
    \end{subfigure}%
    \hfill
    \begin{subfigure}[b]{.45\linewidth}
        \scalebox{.8}{
            \begin{forest}, nice empty nodes
                [\textit{v}p [$\sqrt{~}$\textipa{xSv}] [\textit{v}\\CiCCeC]]
                \coordinate (low1) at (-1.4,0);
                \coordinate (low2) at (-1.4,-2.4);
                \draw [decorate, decoration = {calligraphic brace}] (low2) -- (low1);
                \node[text width=3cm] at (-2.7,-1.1) {\begin{center}
                    \textipa{xiSev}\\($\sqrt{~}$-derived)
                \end{center}};
            \end{forest}
        }
            \caption{Root derived verb}
            \label{fig:root-derived-structure}
    \end{subfigure}
    \caption{The structure of denominal \textit{vs} root-derived verbs}
    \label{fig:denominal-structures}
\end{figure}

\begin{table}[H]
    \centering
    \small
    \begin{tabular}{l|cc|cl|cl|}
        \cline{2-7}
        & \multicolumn{2}{l|}{\backslashbox{Pattern}{Root}} & $\sqrt{\text{\textipa{xSv}}}$ &  & $\sqrt{\text{\textipa{ktv}}}$ &  \\ \cline{2-7} 
        (viii) & \textcolor{blue}{m}i\textbf{CC}a\textbf{C}a & (n) & \textipa{\textcolor{blue}{m}a\textbf{xS}a\textbf{v}a} & `thought' &    \textipa{\textcolor{blue}{m}i\textbf{xt}a\textbf{v}a} & `desk' \\
        (ix) & \textcolor{blue}{m}a\textbf{CC}e\textbf{C} & (n) & \textipa{\textcolor{blue}{m}a\textbf{xS}e\textbf{v}} & `computer' & NA &  \\
        (x) & \textcolor{blue}{m}i\textbf{CC}u\textbf{C} & (n) & \textipa{\textcolor{blue}{m}i\textbf{xS}u\textbf{v}} & `computing' & NA & \\
        (xi) & \textbf{CC}i\textbf{C}u\textcolor{blue}{t}    & (n) & \textipa{\textbf{x}a\textbf{S}i\textbf{v}u\textcolor{blue}{t}} & `importance' & NA &  \\
        (xii) & \textcolor{blue}{t}a\textbf{CC}i\textbf{C} & (n) & \textipa{\textcolor{blue}{t}a\textbf{xS}i\textbf{v}} & `calculation' & \textipa{\textcolor{blue}{t}a\textbf{xt}i\textbf{v}} & `decree' \\
        (xiii) & \textbf{C}e\textbf{CC}o\textcolor{blue}{n} & (n) & \textipa{\textbf{x}e\textbf{Sb}o\textcolor{blue}{n}} & `bill' & NA &  \\
        (xiv) & \textbf{CC}i\textbf{C}a & (n) & \textipa{\textbf{x}a\textbf{S}i\textbf{v}a} & `thinking' & \textipa{\textbf{kt}i\textbf{v}a} & `writing' \\
        (xv) & \textbf{CC}a\textbf{C} & (n) & NA & & \textipa{\textbf{kt}a\textbf{v}} & `hand-writing' \\
        (xvi) & \textbf{C}a\textbf{C}a\textbf{C} & (n) & \textipa{\textbf{x}a\textbf{S}a\textbf{v}} & `accountant' & \textipa{\textbf{k}a\textbf{t}a\textbf{v}} & `correspondent' \\
        (xvii) & \textcolor{blue}{m}i\textbf{CC}a\textbf{C} & (n) & NA    &  & \textipa{\textcolor{blue}{m}i\textbf{xt}a\textbf{v}} & `letter' \\
        (xviii) & \textbf{C}a\textbf{C}u\textbf{C} & (a) & \textipa{\textbf{x}a\textbf{S}u\textbf{v}} & `important' & \textipa{\textbf{k}a\textbf{t}u\textbf{v}} & `written' \\ 
        (xix) & \textbf{C}a\textbf{CC}a\textcolor{blue}{n} & (a) & NA & & \textipa{\textbf{k}a\textbf{tv}a\textcolor{blue}{n}} & `pulp writer'\\\cline{2-7} 
    \end{tabular}
    \caption{Nominal and adjectival patterns}
    \label{fig:nominal-templates}
\end{table}
\vspace{-3mm}
Recall that the two-level model made two predictions: (a) Elements derived from the same \textit{root} \textit{via} a lower-level morphological operation may arbitrarily differ semantically; and (b) elements that derive from the same \textit{word} \textit{via} upper-level morphological operations should be closely related semantically.
The data in Tab. \ref{fig:varieties-of-meaning} and \ref{fig:nominal-templates} seem to confirm the former prediction, as the meanings that result from combining the different roots with nominal and adjectival patterns seem quite unpredictable. Arad \cite{Arad2003} claims that Hebrew provides us with the possibility to test the second prediction. More specifically, Hebrew comes with a morphophonemic diagnostic that can help identify elements derived \textit{via} upper-level operations, such as denominal and de-adjectival verbs.\footnote{In the prose, we will refer going forward only to denominal verbs, but the observations apply to de-adjectival ones as well.} Denominal verbs are derived by merging a verbal pattern with a \textit{noun}, rather than with a \textit{root}. The base noun is itself derived from a root:
\begin{equation*}
    \sqrt{~} \stackrel{lower-level}{\longrightarrow} N \stackrel{upper-level}{\longrightarrow} V_{denom}
\end{equation*}
In Hebrew, certain nominal patterns include templatic consonants (marked in blue in the Figures), in addition to the templatic vowels. For instance, templates (viii-x) and (xvii) in Tab. \ref{fig:nominal-templates} have a templatic /m/, templates (xi-xii) have a templatic /t/, and templates (xiii) and (xix) have a templatic /n/. Arad notes that certain \textit{verbs} (such as \textipa{mixSev} and \textipa{hitxaSben} in Tab. \ref{fig:nouns-to-denominals-table} below) seem to behave as if a consonant has been adjoined to their root. She argues that this consonant is a templatic consonant originating from the root-derived noun that served as a base for the derivation of the corresponding verb (\textit{denominal}). In other words, it is argued that \textipa{mixSev} and \textipa{hitxaSben} do not derive directly form the root $\sqrt{\text{\textipa{xSv}}}$, but rather, from the root-derived nouns \textipa{maxSev} and \textipa{xeSbon}, respectively. This is illustrated in Fig.  \ref{fig:denominal-structure}. Other verbs, such as \textipa{xiSev} and \textipa{hitxaSev}, which are derived using the same pattern as \textipa{mixSev} and \textipa{hitxaSben} respectively (cf. Tab. \ref{fig:nouns-to-denominals-table}), lack a templatic consonant and are therefore argued to derive \textit{directly} from their respective roots. This is illustrated in Fig. \ref{fig:root-derived-structure}. Also note in Tab. \ref{fig:nouns-to-denominals-table} that verbal forms involving a templatic consonant (\textipa{mixSev}, \textipa{hitxaSben}) have a meaning which seems closely related to the meaning of the nouns derived from the same root (\textipa{maxSev}, \textipa{xeSbon}). If the presence of a templatic consonant is indeed the marker of an upper-level operation mapping a noun to a verb, as Arad argues, this semantic property is in line with \textbf{Prediction (b)} of two-level model (cf. p.~3). In contrast, verbal forms derived using the same patterns, but devoid of any templatic consonant (\textipa{xiSev}, \textipa{hitxaSev}) seem not as close to the corresponding root-derived nouns in terms of meaning, this time in line with \textbf{Prediction (a)} of the two-level model.

\begin{table}[H]
    \centering
    \small
    \begin{tabular}{|c|cl|c|cl|}
        \hline
        N pattern & \multicolumn{2}{c|}{$\sqrt{~}$-derived Noun} & V pattern & \multicolumn{2}{c|}{Possible Verbs} \\ \hline
        \textcolor{blue}{m}a\textbf{CC}e\textbf{C} & \textcolor{blue}{m}\textipa{a\textbf{xS}e\textbf{v}} & `computer' & \textbf{C}i\textbf{CC}e\textbf{C} & \textcolor{blue}{m}\textipa{i\textbf{xS}e\textbf{v}} & `computerized'\\
        (ix)& & & & \textipa{\textbf{x}i\textbf{S}e\textbf{v}} & `calculated' \\ \hline
        \textbf{C}e\textbf{CC}o\textcolor{blue}{n} & \textipa{\textbf{x}e\textbf{Sb}o}\textcolor{blue}{n} & `bill' & hit\textbf{C}a\textbf{CC}e\textbf{C} & \textipa{hit\textbf{x}a\textbf{Sb}e}\textcolor{blue}{n} &`settled up financially'\\
        (xiii)& & & & \textipa{hit\textbf{x}a\textbf{S}e\textbf{v}}& `was considerate'\\\hline
    \end{tabular}
    \caption{The root $\sqrt{\text{\textipa{xSv}}}$ and the templatic consonants}\
    \label{fig:nouns-to-denominals-table}
    \vspace{-20pt}
\end{table}
\vspace{-3mm}
\vspace{-1mm}
In brief, the two-level model, together with Arad's claim that verbs with an extra nominal-template consonant are denominal rather than root-derived, predicts that these denominal verbs exhibit a closer semantic similarity to the noun from which they were derived, than root-derived verbs do.

\subsection{Previous empirical investigations into denominals in the two-level model}

As far as we know, our paper is the first to investigate the semantic encoding of Hebrew denominal verbs within word embedding models. However, an experimental study on the processing of Hebrew denominal verbs has already been conducted by Brice on human participants \cite{Brice2016}. A priming experiment was used to provide further evidence for the two-level model, based on the background assumption that morphological inclusion corresponds to a strong priming connection. This claim is supported by a series of papers (\cite{Frost1997}, \cite{Frost2000}, \cite{Frost2005}, \cite{Velan2007}, \cite{Velan2011}), showing that roots have a strong priming effect for verbs that are derived from them. More generally, it can be assumed that if some language component \textit{A} is contained in the morphological structure of a component \textit{B}, then the string that orthographically represents \textit{A} should prime the string that represents \text{B}. In \cite{Brice2016}, Brice tests this property on Hebrew nouns and denominals, by comparing the priming effect of a noun on the denominal verb derived from it, to that of a verb derived from the same root. An example for such triplets is given in Tab. \ref{tab:priming}. Crucially, the two stimuli were orthographically equally similar to the target, so any difference in priming could only be attributed to deeper connections between the words.

\begin{wraptable}[5]{r}{.45\linewidth}
    \centering
    \small
    \vspace{0mm}
    \begin{tabular}{|r|l l|}
        \hline
        Target & \textipa{letaxkeK} & `to debrief' \\ \hline
        Noun stimulus & \textipa{taxkiK} & `debriefing' \\ \hline
        Verb stimulus & \textipa{laxkoK} & `to inquire' \\ \hline
    \end{tabular}
    \caption{Example stimulus from \cite{Brice2016}}
    \label{tab:priming}
\end{wraptable}

Given these assumptions, the two-level model's prediction is that the noun stimulus should have a stronger priming effect on the denominal than the root-derived verb stimulus. That is because the noun is contained in the denominal verb's morphological structure, while the root-derived verb is not. Brice indeed found that the noun stimuli yield significantly shorter reaction times among the participants than the verb stimuli, i.e., demonstrate a stronger priming effect. We view our study as complementing this result. However, Brice argued that a given noun's priming its corresponding denominal verb could not be explained by a semantic connection between them, since priming has been shown to be unaffected by the meaning of the stimulus (see \cite{Perea2002}, \cite{Neely2012}, a.o.). Therefore, Brice took his results as evidence for the morphosyntactic aspect of the two-level model. Our study, on the other hand, aims to provide evidence in favor of the existence of semantic consequences of the the two-level model; i.e., the idea that upper-level derivational operations entail a high degree of semantic similarity between their input and output.
\vspace{-3mm}

\section{Testing the predictions using Hebrew word-embedding models}

\subsection{Words embeddings capture meaningful semantic generalizations}

Static word embedding models such as Word2Vec \cite{Mikolov2013}, GloVe \cite{Pennington2014}  and fastText \cite{Bojanowski2016} map each word of a given lexicon to a dense, high-dimensional vector. This mapping is obtained by training a neural network on specific language-related tasks, such as predicting the context of any given word (Skip-gram), or predicting a word given a context (CBOW) \cite{Mikolov2013}. By contrast, contextualized word embeddings like BERT \cite{Devlin2018}, and its variant AlephBERT \cite{Seker2021}, pretrained on Hebrew data, can take whole sentences as input, and may map the same word to different representations, depending on the context. BERT in particular, adopts a deep encoder-decoder architecture (``Transformer'', \cite{Vaswani2017}). A stack of encoders (12 for the basic model) use attention mechanisms to forward a more complete picture of the whole sequence to the decoder.

Word embeddings in general have been argued to encode lexical meaning, in the sense that word-vectors that are close to each other in the embedding space are expected to be close in meaning \cite{Jurafsky2000}. The relevant metric is usually taken to be the cosine similarity (measure of the angle) between two vectors. In particular, word embeddings have been shown to encode specific morphological/semantic relationships (such as comparative/superlative formation, masculine/feminine nominalizations, some part-whole relationships) as stable linear transformations, a somewhat surprising result that has led to various explanations in the recent years
\cite{Arora2016,Gittens2017,Ethayarajh2018,Allen2019}. Contextualized embeddings like BERT were shown to be especially good in capturing polysemy and homonymy in natural language \cite{Nair2020}. Given that word embeddings provide a quantitative measure of semantic similarity, and that our morphological model makes specific semantic predictions, those language models appear as an interesting testing ground for the predictions of the two-level model.

\subsection{Word embeddings and the two-level model}

We propose here that the abstract linguistic notion of \textit{root} be modeled as a subspace of the word embedding. This region should contain (at least) all the vectors corresponding to words derived from the root through a merger with a functional head (more specifically in our case, a pattern).\footnote{A possible way to define this region in mathematical terms would be to use the notion of convex hull, although our analysis does not depend on any specific implementation thereof.} For instance, the root $\sqrt{\text{\textipa{xSv}}}$ designates a region that contains, among others, the vectors for \textipa{xaSuv} (`important'), \textipa{maxSava} (`thought'), and \textipa{hitxaSev} (`was considerate').

In the two-level framework, the generation of root-derived elements is semantically opaque (cf. \textbf{Prediction (a)}). In other words, elements derived from the same root \textit{via} a lower-level process (i.e. the merger of a head directly with the root), are expected to exhibit arbitrary semantic differences. Assuming that roots denote regions in the embedding space, semantic opacity corresponds to an expectation that for any root $\sqrt{~}_x$, and any set of templates $\{t_1, t_2, \ldots, t_i\}$, the vectors corresponding to the words derived from applying the templates to the root (i.e., \footnotesize $\{\overrightarrow{t_1(\sqrt{~}_x)}, \overrightarrow{t_2(\sqrt{~}_x)}, \ldots, \overrightarrow{t_i(\sqrt{~}_x)}\}$ \normalsize) will be arbitrarily distributed over the region designated by the root.

\begin{wrapfigure}[12]{r}{.4\linewidth}
    \centering
    \vspace{-4mm}
    \includegraphics[width=.9\linewidth]{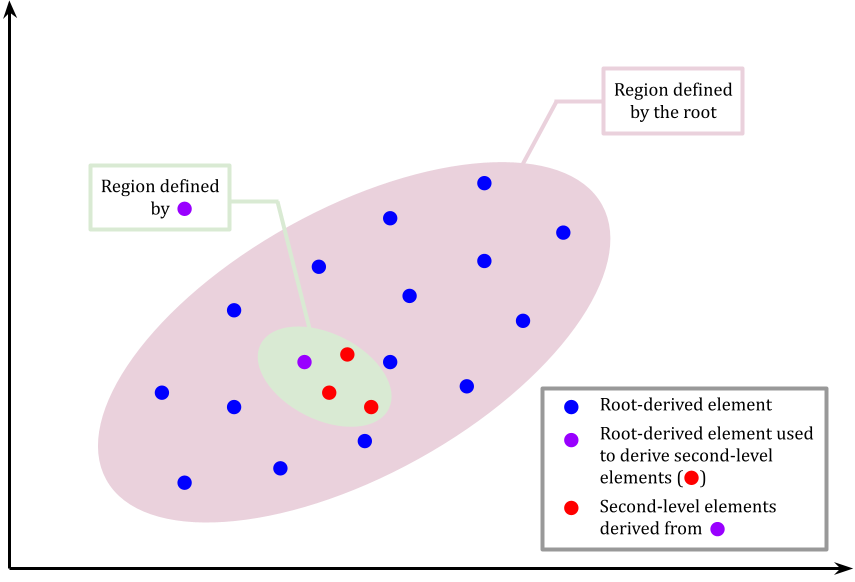}
    \caption{Expected distribution of root-derived \textit{vs} word-derived elements within a simplified embedding space.}
    \label{fig:space-schema}
\end{wrapfigure}

The generation of word-derived elements, on the other hand, is argued to be semantically restricted and transparent (cf. \textbf{Prediction (b)}). More specifically, if $Y$ is an element derived by merging a functional head with an element that already contains a functional head $X$, then the meaning of $Y$ is expected to be close to the meaning of $X$ in a systematic way \cite{Arad2003}. In other words, if the merger of the first functional head can lead to a vector $\vec{X}$ located anywhere within the region denoted by the root (Prediction (a)), the merger of a second head on $X$ (yielding $Y$) should lead to a representation $\vec{Y}$ that is in the close vicinity of $\vec{X}$ (Prediction (b)).  An illustration of this interpretation of the two-level model predictions is provided in Fig. \ref{fig:space-schema}.

\subsection{Testing the two-level model in the context of Hebrew word embeddings}
When it comes to Hebrew denominal verbs, the prediction of our interpretation of the two-level model in the context of word embeddings is the following. Given a root ($\sqrt{~}$); a noun derived from it \textit{via} a lower-level operation ($N_{\sqrt{~}}$); a denominal verb derived from that noun \textit{via} an upper-level operation ($V_{N_{\sqrt{~}}}$); and finally, a verb derived directly from the root \textit{via} a lower-level operation ($V_{\sqrt{~}}$), we expect $\overrightarrow{V_{N_{\sqrt{~}}}}$ to be generally\footnote{We think that it is too restrictive to state that this property should hold \textit{for any given root}, due to the arbitrary character of lower-level morphological operations. Since lower-level operations are assumed to map root-derived elements somewhat randomly within the subspace defined by the root, a root-derived verb might accidentally end up very close to a root-derived noun (meaning, closer to this noun than the noun-derived denominal is). Our modeling however, predicts that this configuration should be rare enough for our inequalities to have some statistical significance.} closer to $\overrightarrow{N_{\sqrt{~}}}$ within the embedding space, than $\overrightarrow{V_{\sqrt{~}}}$ is.

The exact nature of the root-derived verb $V_{\sqrt{~}}$ remains to be fleshed out however.
Indeed, if a given root $\sqrt{~}$ normally yields a single root-derived noun ($N_{\sqrt{~}}$), it can in principle give rise to many different root-derived verbs \footnotesize$ \lbrace V_{\sqrt{~}}^{(1)}, \dots , V_{\sqrt{~}}^{(k)} \rbrace$ \normalsize, some of them being closer to $N_{\sqrt{~}}$ than others. Assuming that \footnotesize$ \lbrace \overrightarrow{V_{\sqrt{~}}^{(1)}}, \dots , \overrightarrow{V_{\sqrt{~}}^{(k)}} \rbrace$ \normalsize are somewhat uniformly distributed across the region defined by $\sqrt{~}$, we expect the mean similarity between $\overrightarrow{N_{\sqrt{~}}}$ and each of the $\overrightarrow{V_{\sqrt{~}}^{(i)}}$ to be lower than the similarity between $\overrightarrow{N_{\sqrt{~}}}$ and the denominal derived from it, $\overrightarrow{V_{N_{\sqrt{~}}}}$. This is formalized in Eq. \ref{eq:mean-ineq} below (where $\mathcal{S}$ stands for cosine similarity):
\begin{center}
    \fbox{\begin{minipage}{.7\linewidth}
    \begin{equation}
        \forall \sqrt{~}: \ \frac{1}{k} \sum_{i = 1}^k\mathcal{S}\left(\overrightarrow{N_{\sqrt{~}}}, \overrightarrow{V_{\sqrt{~}}^{(k)}}\right) \ < \ \mathcal{S}\left(\overrightarrow{N_{\sqrt{~}}}, \overrightarrow{V_{N_{\sqrt{~}}}}\right) \label{eq:mean-ineq}
    \end{equation}
    \begin{center}
        \textbf{Hypothesis 1}
    \end{center}
\end{minipage}}
\end{center}

\vspace{2mm}
Assuming that vectors of root-derived verbs are arbitrarily distributed across the region defined by the root, some root-derived verbs might be accidentally closer to the root-derived noun than the denominal verb derived from that noun. By using the mean similarity between the root-derived noun and the various root-derived verbs, the prediction is rendered compatible with this possibility. However, it might also be worthwhile to test a stronger prediction, according to which the similarity between a noun and a denominal verb derived from it should be greater than the similarity between the noun and the root-derived verb that is \textit{maximally similar} to the noun. This is formalized in Eq. \ref{eq:max-ineq} below.
\begin{center}
    \fbox{\begin{minipage}{.7\linewidth}
    \begin{equation}
        \forall \sqrt{~}: \ \text{max}_{i \in [1, k]}\mathcal{S}\left(\overrightarrow{N_{\sqrt{~}}}, \overrightarrow{V_{\sqrt{~}}^{(k)}}\right) \ < \ \mathcal{S}\left(\overrightarrow{N_{\sqrt{~}}}, \overrightarrow{V_{N_{\sqrt{~}}}}\right) \label{eq:max-ineq}
    \end{equation}
    \begin{center}
        \textbf{Hypothesis 2}
    \end{center}
\end{minipage}}
\end{center}
\section{Implementation and results}

\subsection{Dataset creation}
We generated and tested Hebrew data in order to validate our two hypotheses. Each data point in our dataset contained (1) a noun with a templatic consonant, (2) a denominal verb containing the templatic consonant from the noun, in addition to the three root consonants (cf. Tab. \ref{fig:nouns-to-denominals-table}), and (3) a list of verbs derived directly from the same root as the noun (and thus devoid of templatic consonants).\footnote{It is worth mentioning that a single noun could in certain cases give rise to two data points, when the noun's root happened to be compatible with two denominal templates.} Each data point contained between one and five root-derived verbs, depending on how productively the given root combined with the verbal templates.

\begin{wraptable}[12]{r}{0.5\textwidth}
    \centering
    \vspace{-5mm}
    \begin{tabular}{|c|c|}
        \hline
        Nominal pattern  & Denominal pattern \\ \hline
        \textbf{t}iCCoCet  &   \\
        \textbf{t}iCCoCa &   le\textbf{t}aCCeC \\
        \textbf{t}aCCiC    &   \\\hline
        CeCCo\textbf{n}    &  leCaCCe\textbf{n}, lehitCaCCe\textbf{n}  \\ \hline
        \textbf{m}aCCeC    &   \\
        \textbf{m}iCCeCet    &   le\textbf{m}aCCeC, lehit\textbf{m}aCCeC \\
        \textbf{m}iCCaC   &    \\ \hline
        \textbf{\v{s}}aCCeCet   &  le\textbf{\v{s}}aCCeC, lehi\textbf{\v{s}}taCCeC \\\hline
        CaCaCa\textbf{t}   &  leCaCCe\textbf{t}, lehitCaCCe\textbf{t} \\\hline
    \end{tabular}\vspace{-1mm}
    \caption{Templates used for data generation}
    \label{fig:template-mapping}
\end{wraptable}

A list of nominal patterns containing templatic consonants was constructed using introspection and previous linguistic papers on the subject (\cite{BatEl1994, Arad2003, Brice2016}). Each nominal template was mapped to the verbal template which incorporates the nominal consonant into the verbal form. This is illustrated in Tab. \ref{fig:template-mapping} (partial list). Given that Modern Hebrew lacks vowel marking in the orthography, and therefore involves a high rate of ambiguity (cf. \cite{tomer2012}), we used the infinitival form of the verbal templates, which are not ambiguous with nominal elements in the language even in the absence of vowel markings.

A list of nouns instantiating the relevant nominal templates was generated by matching nouns from a PoS-tagged Hebrew corpus, \textit{The Knesset Meetings Corpus}\footnote{This corpus gathers protocols of sessions in the Israeli parliament between January 2004 and November 2005. The particular archive we used was \texttt{kneset16}.}, against the various nominal templates. Candidate denominal verbs corresponding to each noun could be subsequently created, using the template mapping established in Tab. \ref{fig:template-mapping}. Root-derived verbs were generated using the five infinitival verbal templates of Hebrew, corresponding to the seven inflected templates in Tab. \ref{fig:varieties-of-meaning}. This process generated a list of 1435 potential data points. Given that not all nouns can productively give rise to denominal verbs, most of these were not actual verbs of Hebrew, or did not have a root that could productively combine with other verbal templates. We first eliminated the data points that were obviously not part of the grammar, by checking if the denominal (or any inflected form thereof) could be found in the list of verbs extracted from the Knesset dataset. This first filtering step ruled out 1322 data points, and left us with 113 data points to inspect further. We manually discarded the remaining defective items, ending up with a list of 66 denominal verbs.

\subsection{Word embeddings}
\begin{wraptable}[4]{r}{0.56\textwidth}
    \small
    \centering
    \vspace{-26mm}
    \begin{tabular}{|c|rrrr|}\hline
    Model             & Word2Vec & GloVe & fastText & BERT \\ \hline
    \# vectors &    584 160      & 584 162      & 2 billion & NA\footnotemark \\ \hline
    \begin{tabular}[c]{@{}c@{}}Initial\\ dimension\end{tabular} & 100         &  50/100\footnotemark     & 300 & 768   \\ \hline  
    \begin{tabular}[c]{@{}c@{}}PCA-reduced\\ dimension\end{tabular} & 27         &  28/46     &   50 & 107  \\   \hline
    \end{tabular}
\caption{Characteristics of the models}
\label{tab:models-charac}
\end{wraptable}

\footnotetext{We trained GloVe using the default dimension of 50, then switched to 100 to allow for a better comparison with the other models, which have similar dimensions.}
\footnotetext{As BERT models do not assign words to a fixed vector.}

To convert the words in our dataset into vectors, we used pretrained models from fastText \cite{Grave2018}, and BERT (AlephBERT) \cite{Seker2021}, and trained GloVe and Word2Vec\footnote{For training the Word2Vec model, we used the Skip-Gram architecture.} models on Hebrew Wikipedia dumps. The characteristics of those various embeddings are given in Tab. \ref{tab:models-charac}.

The AlephBERT embedding was obtained by summing the last 4 layers obtained after feeding the model with individual tokenized words, one at a time. Tokenization was performed using a dedicated function from the BERT library. If a given word was represented using several tokens, then, the representation of the word was obtained by averaging the representations of the individual tokens.

Since the dimensions of the embeddings were close to our dataset's total size, we reduced the space using Principal Component Analysis (PCA, \cite{Pearson1901}) prior to computing the similarities and testing the hypotheses. We relied on the Guttman-Kaiser Criterion \cite{Guttman1954} to determine the target dimension for each space. Before computing the similarities, we chose to plot a few data points in a 2D space. For the visualizations to be as meaningful and readable as possible, we used PCA with a cosine kernel on each separate data point. A few such plots are represented in Fig \ref{fig:cosine-dists}.

\begin{figure}[H]
    \centering
    \begin{subfigure}[b]{0.32\textwidth}
        \centering
        \includegraphics[trim=38.1 23.1 0 0,clip,width=1\linewidth]{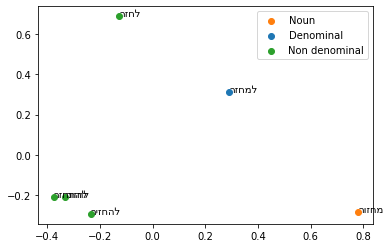}
        \caption{\centering Noun: `cycle';\newline Denominal: `to recycle'}
    \end{subfigure} 
    \hfill
    \begin{subfigure}[b]{0.32\textwidth}
        \centering
        \includegraphics[trim=38.1 23.1 0 0,clip,width=1\linewidth]{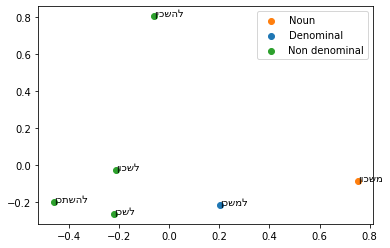}
        \caption{\centering Noun: `pawning';\newline Denominal: `to pawn'}
    \end{subfigure}
    \hfill
    \begin{subfigure}[b]{0.32\textwidth}
        \centering
        \includegraphics[trim=38.1 23.1 0 0,clip,width=1\linewidth]{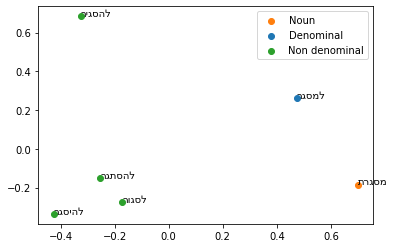}
        \caption{\centering Noun: `frame';\newline Denominal: `to frame'}
    \end{subfigure}
\end{figure}%
\vspace{-7mm}
\begin{figure}[H]\ContinuedFloat
    \centering
    \begin{subfigure}[b]{0.4\textwidth}
        \centering
        \includegraphics[trim=38.1 23.1 0 0,clip, width=.8\linewidth]{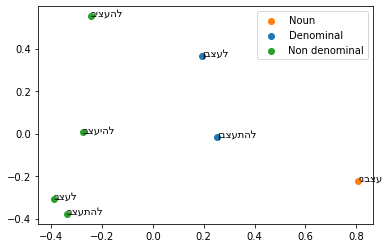}
        \caption{\centering Noun: `annoyed'\newline Denominals: `to get annoyed', `to annoy'}
    \end{subfigure}
    \qquad
    \begin{subfigure}[b]{0.4\textwidth}
        \centering
        \includegraphics[trim=38.1 23.1 0 0,clip,width=.8\linewidth]{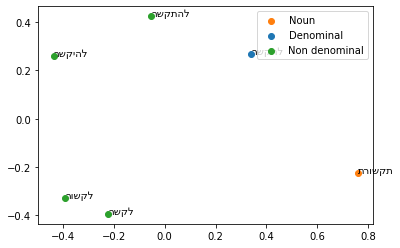}
        \caption{\centering Noun: `communication';\newline Denominal: `to communicate'}
    \end{subfigure}
    \caption{2D projection of a few data points (PCA, cosine kernel, fastText model)}
    \label{fig:cosine-dists}
\end{figure}

\noindent Those 2D plots bring preliminary evidence in support of the denominal verbs (blue dots) being closer to their respective nouns (orange dots) in a cosine space, than the other root-derived verbs (green dots). 
\subsection{Computation of the similarities}
For each data point, our main hypothesis predicts that the cosine similarity between the noun and the denominal verb should be higher than the mean cosine similarity between the noun and all the other verbs sharing the same root. The stronger hypothesis predicts that the same kind of inequality holds when the ``mean'' operator is replaced by a ``max'' operator. In both cases, for each data point, the difference between two measures of similarities (noun/denominal; noun/other root derived element) is expected to be positive. The distributions of those two measures of similarity are represented in Fig. \ref{fig:sims-diffs}, for the main hypothesis, and the various models we tested. Those plots suggest that the main hypothesis is verified, since all distributions seem to have a mean and median above 0. In the next section, we test the significance of those empirical observations.

\begin{figure}[H]
    \centering
    \begin{subfigure}[b]{0.32\textwidth}
        \centering
        \includegraphics[width=\linewidth]{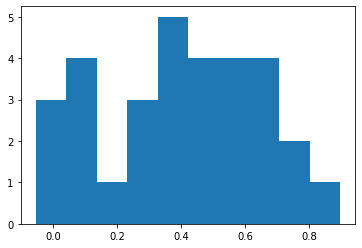}
        \caption{\centering Word2Vec}
        \label{fig:sims-mean}
    \end{subfigure}
    \hfill
    \begin{subfigure}[b]{0.32\textwidth}
        \centering
        \includegraphics[width=\linewidth]{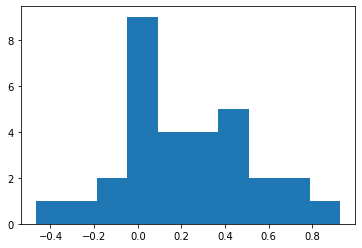}
        \caption{\centering GloVe50}
        \label{fig:diffs-mean}
    \end{subfigure}
    \hfill
    \begin{subfigure}[b]{0.32\textwidth}
        \centering
        \includegraphics[width=\linewidth]{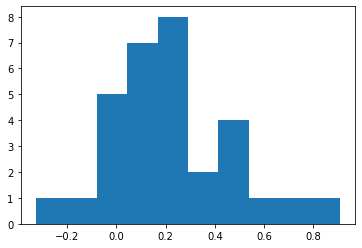}
        \caption{\centering GloVe100}
        \label{fig:sims-best}
    \end{subfigure}
\end{figure}%
\vspace{-7mm}
\begin{figure}[H]\ContinuedFloat
    \centering
    \begin{subfigure}[b]{0.4\textwidth}
        \centering
        \includegraphics[width=.8\linewidth]{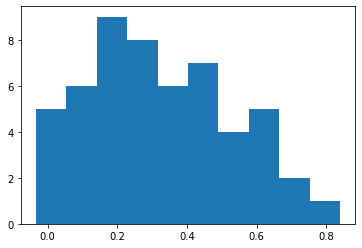}
        \caption{\centering fastText}
        \label{fig:diffs-best}
    \end{subfigure}
    \qquad
    \begin{subfigure}[b]{0.4\textwidth}
        \centering
        \includegraphics[width=.8\linewidth]{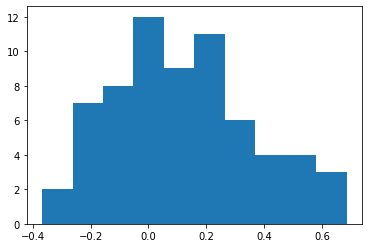}
        \caption{\centering AlephBERT}
        \label{fig:diffs-mean}
    \end{subfigure}
    \caption{Paired similarity differences for various models under H1 (main hypothesis)}
    \label{fig:sims-diffs}
\end{figure}

\subsection{Tests and results}
To test whether denominals were significantly more similar to their corresponding nouns than other root-derived verbs, we performed (non-parametric) one-tailed Wilcoxon tests for matched-pairs\footnote{We preferred a non-parametric test as opposed to a standard t-test because the similarity plots suggested that the distributions did not satisfy the t-test assumptions. This was confirmed by Levene tests conducted prior to performing the main tests, for all models but AlephBERT.} on the data described and plotted in the previous section.  We did not feel the need perform any correction on the $p$-values, even though 2 hypotheses were tested for each model, because H1 was entailed by H2 by construction (so, a significant $p$-value for H2 implies a significant $p$-value for H1 as well). The $p$-values and effect sizes are compiled in Tab. \ref{tab:test-results}. The effect sizes correspond to Cliff's $\Delta$ \cite{Cliff1993}, which is a robust measure for non-parametric samples.
\begin{table}[H]
    \centering
    \begin{tabular}{|c|r|r|r|r|r|}
        \hline  & Word2Vec & GloVe50 & GloVe100 & fastText & AlephBERT\\ \hline
        \# data points  & 31 & 31 & 31 & 53 & 66\\ \hline
        H1 &\cellcolor[HTML]{9AFF99}\begin{tabular}[c]{@{}r@{}}$1.06\times 10^{-6}$\\ 0.86 (Large)\end{tabular}& \cellcolor[HTML]{9AFF99}\begin{tabular}[c]{@{}r@{}}$2.43 \times 10^{-4}$\\ 0.52 (Large)\end{tabular}&  \cellcolor[HTML]{9AFF99}\begin{tabular}[c]{@{}r@{}}$6.64 \times 10^{-5}$\\ 0.66 (Large)\end{tabular} &  \cellcolor[HTML]{9AFF99}\begin{tabular}[c]{@{}r@{}}$1.42\times 10^{-10}$\\ 0.79 (Large)\end{tabular} &  \cellcolor[HTML]{9AFF99}\begin{tabular}[c]{@{}r@{}}$4.84\times 10^{-4}$\\ 0.30 (Small)\end{tabular}\\ \hline
        H2 & \cellcolor[HTML]{9AFF99}\begin{tabular}[c]{@{}r@{}}$3.77 \times 10^{-5}$\\ 0.66 (Large)\end{tabular}&\cellcolor[HTML]{FD6864} \begin{tabular}[c]{@{}r@{}} $1.68 \times 10^{-1}$\\ 0.06 (Negligible)\end{tabular} & \cellcolor[HTML]{9AFF99}\begin{tabular}[c]{@{}r@{}}\cellcolor[HTML]{9AFF99}$2.87 \times 10^{-2}$\\ 0.20 (Small)\end{tabular} & \cellcolor[HTML]{9AFF99} \begin{tabular}[c]{@{}r@{}}$1.39\times 10^{-8}$\\ 0.62 (Large)\end{tabular} &  \cellcolor[HTML]{FD6864}\begin{tabular}[c]{@{}r@{}}$3.59\times 10^{-1}$\\ 0.02 (Negligible)\end{tabular}\\ \hline
    \end{tabular}
    \caption{$p$-values and effect sizes (Cliff's $\Delta$) for H1 and H2 and 4 embedding models}
    \label{tab:test-results}
\end{table}

\vspace{-5mm}
As shown above, our main hypothesis (H1) was verified in all the models tested, and the stronger hypothesis (H2) appeared to be verified in all but two models (GloVe50 and AlephBERT). This indicates that our main prediction is quite robust across various language models. Overall large effect sizes also suggest that the effect, whenever present, is quite strong.

The fact that GloVe100, but not GloVe50, verified both hypotheses is somewhat interesting, as this might mean that a certain richness is needed in the original embedding space (despite the fact that this space is subsequently reduced \textit{via} PCA), in order to capture the relevant generalization. This claim however, seems to be disproved by the results of AlephBERT, since this model failed to verify H2 just like GloVe50, while having the highest original dimension (768). Moreover, the failure of AlephBERT on H2 does not seem to be caused by the dimension reduction process (PCA) being too permissive in keeping too many irrelevant dimensions. Indeed, reducing the space further to the arbitrary dimension of 50 (which retained 71\% of the explained variance) did not change the overall outcome -- H2 was still rejected. Rather, the inefficiency of AlephBERT may be caused by the relative misuse of that model in the context of our experiment; sequential models such as BERT perform well on contextualized data, but here, the model was fed with isolated words, and could not benefit from the presence of surrounding words to enrich its representations.

\section{Discussion, conclusion, and further questions}
Our results seem to corroborate the claim of the two-level model that word-derived elements are systematically closer semantically to the word from which they derive, than elements derived from the same root are similar to that word. This prediction was verified in all the word embeddings we tested, provided that the original dimension was high enough. This result complements Brice's contribution (cf. \cite{Brice2016}), by establishing that morphological inclusion has some real semantic import, in addition to having consequences in terms of priming. It also gives support to the claim that word embedding models might not only capture superficial linguistic regularities; and that, at least in some very specific corners of the language, those models show behaviors similar to humans'.

However, there is one potential confound in the way we tested our prediction. Static word embedding models obtain a single global representation for each word \cite{Liu2020}. Therefore, those models fail to capture the different meanings of ambiguous words. As mentioned earlier, the absence of vowel marking in Modern Hebrew orthography (coupled with the ``double life'' of certain letters that have multiple pronunciations) renders many written words in Hebrew highly ambiguous. When it comes to denominal verbs matters get even worse, as certain nominal templates that give rise to such verbs are systematically ambiguous with certain inflections of a corresponding root derived verb or with an inflection of the denominal verb itself. The systematically ambiguous templates are listed in Tab. \ref{fig:systematically-ambiguous-patterns}.

\begin{table}[H]
    \centering
    \begin{tabular}{|l|l|ll|}
    \hline
    Templates\footnotemark & Possible PoS \& Morphological features      & \multicolumn{2}{l|}{Example} \\ \hline
    \multirow{2}{*}{\textcolor{blue}{t}(a)\textbf{CC}i\textbf{C}} & Noun with templatic consonant               & \textipa{taklit}    & `record'\\
    & \textsc{2.M.SG Imperative} \textbf{root-derived} verb in hi\textbf{CC}i\textbf{C} & \textipa{taklit}    & `record!' \\ \hline
    \multirow{3}{*}{\begin{tabular}[c]{@{}l@{}}\textcolor{blue}{m}(a)\textbf{CC}(e)\textbf{C},\\ \textcolor{blue}{m}(i)\textbf{CC}(a)\textbf{C}\end{tabular}}     & Noun with templatic consonant               &  \textipa{max\v{s}ev} & `computer'      \\
    & \textsc{3.M.SG Present} \textbf{root-derived} verb in \textbf{C}i\textbf{CC}e\textbf{C}    & \textipa{mexa\v{s}ev}          & `calculates'     \\
    & \textsc{3.M.SG Past} \textbf{denominal} verb in \textbf{C}i\textbf{CC}e\textbf{C}          &      \textipa{mix\v{s}ev}     & `computerized'   \\ \hline
    \multirow{2}{*}{\begin{tabular}[c]{@{}l@{}}\textbf{C}(a)\textbf{CC}(a)\textcolor{blue}{n},\\ \textbf{C}(a)\textbf{C}(a)\textbf{C}(a)\textcolor{blue}{t},\end{tabular}} & Noun with templatic consonant               &    \textipa{\texttslig alaxat}       & `plate'          \\
    & \textsc{3.M.SG Past} \textbf{denominal} verb in \textbf{C}i\textbf{CC}e\textbf{C}          &    \textipa{\texttslig ilxet}       & `plated'         \\ \hline
    \end{tabular}
    \caption{Systematically ambiguous patterns}
    \label{fig:systematically-ambiguous-patterns}
\end{table}

\footnotetext{Vowels that are not orthographically represented are put in parentheses.}

\vspace{-3mm}
This makes testing predictions regarding Hebrew in a static word embedding model problematic to begin with, and even more so when it comes to predictions regarding Hebrew denominal verbs. One way to minimize ambiguity in our data was to use forms of each word that are least ambiguous. For instance, to resolve the ambiguity in (i), we used the plural inflections of nouns in the taCCiC template, which are no longer ambiguous with any root-derived verbs. This fix, however, does not resolve the ambiguity in (ii), as the plural form of the noun in that case is still ambiguous -- this time with a plural root-derived verb. We also tried to minimize ambiguity by using verbs in their infinitival form, as infinitival verbs are not ambiguous with any nouns. However, this does not fully resolve the confound, as the infinitival form of verbs in Hebrew involves concatenation of the prefix /\textipa{le-}/ to some combination of the root consonants with other templatic information, and the prefix /\textipa{le-}/ itself is ambiguous between an infinitival marker and the preposition `to'. Therefore, the orthographic representation of some of our denominal verbs is ambiguous between a verbal interpretation and a the prepositional interpretation `to N', where N is the noun from which the denominal is derived. It is unclear to us how the possible ambiguities discussed here influenced our results.

Another way to avoid the problems imposed on us by the high ambiguity rates in Modern Hebrew could be to test our hypotheses in a proper contextual word embedding, obtained by feeding AlephBERT with words put in disambiguating contexts.  However, choosing this solution raises the issue of the design of a relevant context for each target word. First, choosing the right context for a given word is a subjective task that might make our experimental set up biased, or, at least, less controlled. Second, verbs derived from the same root but associated to different templates have different valences and senses, and require complements of different kinds. Those fundamental structural differences would add extra noise that would influence the similarity comparisons, and could not be reasonably counterbalanced. We have yet to investigate how to use a proper contextual embedding model while overcoming this concern.

\newpage
\bibliographystyle{eptcs}
\bibliography{bibliography}
\end{document}